\documentclass{article}

\PassOptionsToPackage{numbers, compress}{natbib}

\usepackage[preprint]{neurips_2024}

\usepackage[utf8]{inputenc} % allow utf-8 input
\usepackage[T1]{fontenc}    % use 8-bit T1 fonts
\usepackage{aliascnt}
\usepackage{hyperref}       % hyperlinks
\usepackage{url}            % simple URL typesetting
\usepackage{booktabs}       % professional-quality tables
\usepackage{amsfonts}       % blackboard math symbols
\usepackage{nicefrac}       % compact symbols for 1/2, etc.
\usepackage{microtype}      % microtypography
\usepackage{xcolor}         % colors

\usepackage{graphicx}
\usepackage{subfigure}
\usepackage{booktabs} % for professional tables
\usepackage{multirow}
\usepackage{amsmath}
\usepackage{amssymb}
\usepackage{mathtools}
\usepackage{amsthm}
\usepackage{bm}
\newaliascnt{suppfigure}{figure}

\newcommand{\beginsupplement}{%
        \setcounter{table}{0}
        \renewcommand{\thetable}{S\arabic{table}}%
        \counterwithin*{figure}{section}
        \renewcommand{\thefigure}{S\arabic{figure}}%
     }

\usepackage[capitalize,noabbrev]{cleveref}

\theoremstyle{plain}

\theoremstyle{definition}

\theoremstyle{remark}

\usepackage[textsize=tiny]{todonotes}

\title{Assessment of Uncertainty Quantification in Universal Differential Equations}

\author{%
Nina Schmid$^{1}$ \quad David Fernandes del Pozo$^{2}$ \quad Willem Waegeman$^{2,*}$ \quad Jan Hasenauer$^{1,3,*}$ \\[0.4em]
\\
$^1$ Life \& Medical Sciences (LIMES) Institute, \\ University of Bonn, Bonn, Germany \\[0.4em]
$^2$ Department of Data Analysis and Mathematical Modelling, \\ Ghent University, Ghent, Germany \\[0.4em]
$^3$ Helmholtz Center Munich, German Research Center for Environmental Health, \\ Computational Health Center, Munich, Germany
\\
\\
$^{*}$ shared last authorship \\
\\
\texttt{nina.schmid@uni-bonn.de}\\
}

\begin{document}

\maketitle

\begin{abstract}
Scientific Machine Learning is a new class of approaches that integrate physical knowledge and mechanistic models with data-driven techniques for uncovering governing equations of complex processes. Among the available approaches, Universal Differential Equations (UDEs) are used to combine prior knowledge in the form of mechanistic formulations with universal function approximators, like neural networks. 
Integral to the efficacy of UDEs is the joint estimation of parameters within mechanistic formulations and the universal function approximators using empirical data.
The robustness and applicability of resultant models, however, hinge upon the rigorous quantification of uncertainties associated with these parameters, as well as the predictive capabilities of the overall model or its constituent components. With this work, we provide a formalisation of uncertainty quantification (UQ) for UDEs and investigate important frequentist and Bayesian methods. By analysing three synthetic examples of varying complexity, we evaluate the validity and efficiency of ensembles, variational inference and Markov chain Monte Carlo sampling as epistemic UQ methods for UDEs.
\end{abstract}

\section{Introduction}
Two primary paradigms govern the modelling of dynamical systems: Mechanistic modelling relies on first principles translated into context-specific formulations \cite{Kitano2002}, while machine learning constructs models through purely data-driven approaches. Scientific machine learning (SciML) unites these paradigms \cite{osti_1478744}, with Universal Differential Equations (UDEs) \cite{rackauckasUniversalDifferentialEquations2021} standing out as a representative example. UDEs describe the dynamics of a process by parameterizing the time-derivative of its state variables. The parametrization is based on blending mechanistic terms with universal function approximators like neural networks, which capture unknown phenomena. With this, UDEs differ from many other SciML methods, like for example Physics-Informed Neural Networks (PINNs) \cite{distinction_scientific_methods}. The parametrization of UDEs allows for the formulation of a variety of hard constraints like mass conservation or boundedness and, hence, enables better model generalization. 

The parameters of both the mechanistic and the neural network components of a UDE are jointly estimated using data. Interpretation of modelling results hinges on quantifying uncertainties, encompassing mechanistic parameter values and predictions for the entire model or its components. 
UQ of parameters is crucial because it provides insights into the reliability and range of potential values, allowing researchers to understand the robustness and credibility of their model's mechanistic foundations. UQ of the prediction is equally important, as it offers a measure of the model's reliability, e.g. for perturbation studies or scenario analysis, aiding decision-making by acknowledging the inherent uncertainty in forecasting outcomes. %Together, the quantification of both parameter and prediction uncertainty enhances the overall interpretability and trustworthiness of modelling results, fostering informed decision-making in various applications.

Uncertainty quantification is a highly researched topic for both dynamical mechanistic modelling \cite{villaverde_assessment_2023} and machine learning \cite{hullermeier_aleatoric_2021,ABDAR2021243}. First-order uncertainty describes the inherent and irreducible stochasticity of the predictions (aleatoric uncertainty), while second-order uncertainty describes uncertainty originating from the uncertainty of parameter estimates (which is one component of epistemic uncertainty -- see \cref{sec:sources of uncertainty}). By choosing a suitable noise model, the assessment of aleatoric uncertainty is well-defined. Hence, in this work, while reporting results on aleatoric uncertainty, we focus on estimating epistemic uncertainty.

In supervised machine learning, various methods aim to quantify epistemic uncertainty, and many of these methods have a resemblance in the field of mechanistic modelling. A fully Bayesian perspective is realised by Markov-Chain-Monte-Carlo (MCMC) sampling methods \cite{MR0031341} and approximation methods like Variational Inference \cite{Jaakkola2000}, yielding parameter distributions instead of point estimates. Deep ensembles \cite{lakshminarayanan_simple_2017} and multi-start ensembles in dynamic modelling \cite{VILLAVERDE201517, villaverde_assessment_2023} are both randomization-based ensemble approaches. Key differences between mechanistic modelling and machine learning are the number and interpretability of the parameters. Hence, some flavours of uncertainty quantification methods are exclusively used in deep learning, like dropout as a Bayesian approximation \cite{pmlr-v48-gal16}. Others are more common in dynamic modelling, like Profile Likelihood (PL) calculation \cite{kreutz_profile_2013} or asymptotic confidence intervals via the Fisher Information Matrix (FIM) \cite{JACQUEZ1985201}. 

For some modelling approaches in the field of SciML, like PINNs \cite{RAISSI2019686}, a thorough investigation of UQ exists \cite{psaros_uncertainty_2023}. In contrast to other methods, UDEs embed neural networks directly in the differential equations. While this allows the incorporation of arbitrary levels of prior knowledge, it also yields unique challenges like over-parametrized differential equations with correlated parameters in combination with numerically more challenging simulations. To the best of our knowledge, previous work explored only basic UQ implementations for multi-start-optimization \cite{akhare2023diffhybriduq} or Bayesian Neural Networks \cite{osti_1888443, dandekar2022bayesian} and only considered fully observed or densely measured state variables. 

In this paper, we present several key contributions. Firstly, we introduce a formal definition of uncertainty tailored to UDEs, aiming to enhance precision and applicability in uncertainty assessments within this framework (\cref{sec:2}). Secondly, we conduct an in-depth discussion of current epistemic UQ methods applicable to UDEs (\cref{sec:epistemic}). Lastly, we evaluate and compare the performance of a diverse set of UQ methods by investigating three synthetic examples (\cref{sec:4}). Each synthetic example is implemented using several noise models, yielding 10 data scenarios in total. Synthetically generated data allows us to compare the methods' results with an underlying ground truth. Our investigation spans considerations of computing time, estimations of aleatoric and epistemic uncertainty, parameter and prediction uncertainty, and different noise models, encompassing continuous and discrete distributions. 
Although the groups of UQ methods have been investigated before \cite{gawlikowski2022survey, villaverde_assessment_2023}, we find novel insights in the context of UDEs.

\section{Formalizing precision: a tailored definition of uncertainty for UDEs}
\label{sec:2}
In the following subsections, we first define the general setup of dynamic models, formally introduce UDEs and conclude by presenting different sources of uncertainty and discussing its relevance for UDEs.

\subsection{Dynamic models}
Let $\bm{x}(t) \in \mathbb{R}^{n_x}$ be a time-dependent variable, denoting the state of a system at time $t$, that can be represented using a dynamic model. Dynamic models describe the value of $\bm{x}$ by parameterizing the derivative of $\bm{x}(t)$ and its initial condition $\bm{x}(t_0)$ using a vector field $f: \mathbb{R} \times \mathbb{R}^{n_x} \times \mathbb{R}^{n_\theta} \to \mathbb{R}^{n_x}$: 
\begin{align}
\label{equ:dynamic_model}
\begin{split}
\frac{d\bm{x}}{dt} &= f(t, \bm{x}, \bm{\theta}_f), \quad \bm{x}(t_0) = \bm{x}_0, \\
\end{split}
\end{align}
where $\bm{\theta}_f \in \mathbb{R}^{n_\theta}$ and $\bm{x}_0 \in \mathbb{R}^{n_x}$ are model parameters and initial conditions, respectively.
Often, $f$ is unknown and an estimate $\hat{f}$ is used instead. Let $\hat{ \bm{\theta}}_f$ and $\hat{\bm{x}}_0$ be the parameters and initial conditions of $\hat{f}$ which we estimate based on $n_t$ discrete measurements at time points $\{t_1, t_2, ..., t_{n_t}\}$. 

 In many real-life scenarios, the state variables cannot be measured directly. Accordingly, the prediction of the differential equation model needs to be transformed using an observable function $h$ to values predicting the measurable observables as:
\begin{align}
\label{eq:combined_de}
\begin{split}
\hat{\bm{y}}(t) &= h(\hat{\bm{x}}(t)) \in \mathbb{R}^{n_y} \quad
\text{with} \quad\hat{\bm{x}}(t) =  \int_{t_0}^{t} \hat{f}(s, \hat{\bm{x}}(s), \hat{\bm{\theta}}_f) \, ds + \hat{\bm{x}}_0.
\end{split}
\end{align}
Here, $\hat{\bm{x}}(t)$ is the estimate of $\bm{x}(t)$ that we get from using $\hat{f}$, $\hat{\bm{\theta}}_f$ and $\hat{\bm{x}}_0$.
An example for $h$ comes from infectious disease modelling, where we often observe infections, but, e.g., not the number of susceptible, exposed or recovered persons. % In this scenario, $\bm{x}(t) = (s(t), e(t), i(t), r(t))^T \in \mathbb{R}^4$ may be a vector describing the number of susceptible $s(t)$, exposed $e(t)$, infected $i(t)$ and recovered persons $r(t)$ of a population at time $t$, and $h: \mathbb{R}^4 \to \mathbb{R}^1; \bm{x}(t) \mapsto i(t)$. 

Furthermore, measurements are subject to noise. Instead of measuring the underlying true value $\bm{\bar{y}}(t_k) = h(\bm{x}(t_k))$ of the observable, we observe the random variable $
\bm{y}(t_k) \sim P$
with $P$ being a probability distribution. In general, we do not know the underlying distribution of $\bm{y}(t_k)$. Instead, we fit a parametric distribution, which is called the noise model. There exist different formulations for noise models, with the Gaussian being the most prominent representative. Depending on the characteristics of the underlying measurements like discreteness, overdispersion, or skewness, other noise formulations may be more suitable. In the present work, we will focus on two commonly used noise models in the context of infectious disease modelling \cite{CONTENTO2023100681, CHAN2021125460}, the Gaussian noise model for continuous data and the Negative Binomial noise model for overdispersed and discrete data: 

\begin{itemize}
\label{sec:noise_model}
\item \textbf{Gaussian noise model}:
Let $\bm{\epsilon}(t_k) \sim \mathcal{N}(\bm{0}, \sigma^2 I)$. Then, we observe $\bm{y}(t_k) = \bm{\bar{y}}(t_k) + \epsilon(t_k)$, where $\sigma$ is the constant standard deviation of the Gaussian distribution. 

\item\textbf{Negative Binomial noise model}:
Let $\bm{y} \in \mathbb{R}^{n_y}$. The observed variable $y_i$ follows a Negative Binomial distribution with mean $\bar{y}_i(t_k)$ and dispersion parameter $d$, i.e.
$y_i(t_k) \sim \mathrm{NegBin}(\bar{y}_i(t_k), d)$, for all $i \in \{1,...,n_y \}$.

\end{itemize}

In both cases, we assume that the i.i.d.\ assumption holds.
Let $ \bm{\theta} = \{\bm{\theta}_f, \theta_{\text{np}} \}$, where $\theta_{\text{np}}$ is the noise parameter of the respective noise model and $p(\bm{y}(t)|\bm{\theta})$  the probability density function with mean value $\hat{\bm{y}}(t)$. Then, the objective of the optimization process is to maximize the likelihood of observing the data $
    \mathcal{D} = \{(t_i, \bm{y}(t_i) | i=1,..., n_t\}$
given the parameters $\bm{\theta}$. % is 
%\begin{align}
%p(\mathcal{D}|\bm{\theta}) = \prod_{i=1}^{n_t} %p(\bm{y}(t_i)|\bm{\theta}).
%\end{align}
 % The parameters of the model are estimated jointly using maximum likelihood-based optimization algorithms. For numerical stability, instead of maximizing the likelihood, the negative log-likelihood is commonly minimized.

\subsection{Universal differential equations} 
UDEs combine known mechanistic terms $f_{\text{mech}}$ with universal function approximators (in this work neural networks) $f_{\text{net}}$ to describe the right-hand side of Eq.~\ref{equ:dynamic_model} \cite{rackauckasUniversalDifferentialEquations2021}. 
For instance, the neural network can be used to describe the time-varying input of an otherwise purely mechanistic ordinary differential equation, i.e. for a fixed $t$ we have $
\hat{f}(t,\bm{x},\bm{\theta}) = \hat{f}_{\text{mech}}(t,\bm{x},\bm{\theta}_f) $, with $\quad \bm{\theta}_f = (\bm{\theta}_{\text{mech}}, \hat{f}_{\text{net}}(t,\bm{\theta}_{\text{net}}))$. Alternatively, it can describe individual terms of the state derivatives, e.g.,
$ \hat{f}(t,\bm{x},\theta) = \hat{f}_{\text{mech}}(t,\bm{x}, \bm{\theta}_{\text{mech}}) + \hat{f}_{\text{net}}(t,\bm{x}, \bm{\theta}_{\text{net}})$.
Hence, the formulation of UDEs allows us to incorporate arbitrary levels of mechanistic knowledge. Here, $\bm{\theta}_{\text{net}}$ are the weights and biases of the neural network and $\bm{\theta}_{\text{mech}}$ are the interpretable parameters of the mechanistic equation. 
Considering all parameters, we define $\bm{\theta} = (\bm{\theta}_{\text{mech}}, \bm{\theta}_{\text{net}}, \bm{\theta}_{\text{np}} )$ for scenarios in which the initial condition $\bm{x}_0$ is known. The parameters are jointly estimated from data.

%While there exist approaches to train UDEs on approximations of the dynamics of a system directly \cite{roesch_collocation_2021}, it is more common to calculate the loss on the observable space. 

\subsection{Sources of uncertainty}
\label{sec:sources of uncertainty}
In general, we can (at least) formally identify two distinct types of uncertainty: \textit{aleatoric} and \textit{epistemic} uncertainty \cite{hullermeier_aleatoric_2021}. As they can guide the evaluation of model performance and its potential application to real-life scenarios, precise quantification of these types of uncertainty is essential. 
The aleatoric (statistical) uncertainty $Var(\bm{y}(t))$ is based on inherent random effects and, hence, irreducible. By introducing a noise model, we aim to describe the aleatoric uncertainty of the model. Epistemic (systematic) uncertainty stems from a lack of knowledge and potential model misspecifications.
The bias-variance decomposition of the mean squared prediction error illustrates these different types of uncertainties \cite{gruber2023sources}:
\begin{align}
\label{equ:bias-variance decomposition}
\begin{split}
\mathbb{E}_{\bm{y}(t)} [ \mathbb{E}_{\mathcal{D}} [ (\bm{y}(t) -\hat{\bm{y}}(t))^2 ]]  % &=  \mathbb{E}_{\bm{y}(t)} [ \mathbb{E}_{\mathcal{D}} [ \bm{y}(t)^2 -2\bm{y}(t)\hat{\bm{y}}(t) + \hat{\bm{y}}(t)^2 ] ] \\ % &= \bm{\bar{y}}(t)^2 + Var(\bm{y}(t)) - 2 \bm{\bar{y}}(t) \mathbb{E}_{\mathcal{D}}[\hat{\bm{y}}(t)] \\ % &+ Var_\mathcal{D}(\hat{\bm{y}}(t))+ (\mathbb{E}_{\mathcal{D}}[\hat{\bm{y}}(t)])^2 \\ 
&= (\mathbb{E}_{\mathcal{D}}[\hat{\bm{y}}(t)] - \bm{\bar{y}}(t))^2 + Var_\mathcal{D}(\hat{\bm{y}}(t)) + Var(\bm{y}(t)) \\
&= bias^2 + Var_\mathcal{D}(\hat{\bm{y}}(t)) + Var(\bm{y}(t)).
\end{split}
\end{align}
Hence, epistemic uncertainty can be decomposed further into model bias $(\mathbb{E}_{\mathcal{D}}[\hat{\bm{y}}(t)] - \bm{\bar{y}}(t))$ and variance $Var_\mathcal{D}(\hat{\bm{y}}(t))$. As described in the previous section, we generally do not know $f$ and $\bm{\theta}$. Uncertainty in the estimates $\hat{\bm{\theta}}$ (model estimation) and $\hat{f}$ (model form), are sources of epistemic uncertainty. 
There exist various methods for the estimation of epistemic uncertainty, as discussed in \cref{sec:epistemic}. However, the model bias is often neglected by assuming $\mathbb{E}_{\mathcal{D}}[\hat{\bm{y}}(t)] = \bm{\bar{y}}(t)$ and reducing the epistemic uncertainty to approximation uncertainty. While we will follow this assumption, we cannot guarantee a negligible model uncertainty: SciML is typically applied to the low to medium data regime \cite{peng_multiscale_2021}. Although neural networks are universal approximators, making them asymptotically unbiased, a bias is typically still observed in the low to medium data regime \cite{belkin_2019,neal_2019_modern}. 

UDEs are located at the interface of neural networks and mechanistic dynamical modelling. While regularisation is vital for neural networks \cite{Goodfellow-et-al-2016}, so is the exhaustive exploration of the parameter space for mechanistic models where one is interested in a global solution. It is not trivial to find the right balance between these, which is one of the reasons why parameter uncertainty, i.e. estimation uncertainty, is of quite some importance for UDEs. Furthermore, the numerical precision of the ODE solver and data sparsity may influence the quality of parameter estimation. 

\section{Methodology for epistemic uncertainty quantification of UDEs}
\label{sec:epistemic}

\subsection{General setting}
In this study, we will explore epistemic uncertainty arising as a result of parameter uncertainty, keeping the model form $f$ fixed per problem setting. 
Bayes' rule provides a formulation for this uncertainty. The posterior density $p(\bm{\theta}|\mathcal{D})$ can be described in terms of the likelihood $p(\mathcal{D}|\bm{\theta})$ and prior $p(\bm{\theta})$:
\begin{align}
\label{equ:Bayes}
\begin{split}
p(\bm{\theta}|\mathcal{D}) = \frac{p(\mathcal{D}|\bm{\theta})p(\bm{\theta})}{\int p(\mathcal{D}|\bm{\theta})p(\bm{\theta}) d\bm{\theta}}.
\end{split}
\end{align}
Using Bayesian model averaging, we obtain the posterior predictive distribution
\begin{align}
\label{equ:posterior_predictive}
\begin{split}
p(\bm{y}(t)|\mathcal{D}) = \int p(\bm{y}(t)|\bm{\theta}) p(\bm{\theta}|\mathcal{D}) d\bm{\theta}.
\end{split}
\end{align}
Bearing in mind that neural network parameters have no physical interpretation, the choice of a prior distribution for its parameters is not trivial. Commonly, an isotropic Gaussian prior is chosen \cite{pml2Book}. Recently, it has been shown that especially for deep and flexible neural networks, this can cause drawbacks like the cold-posterior effect \cite{pmlr-v119-wenzel20a}. Specifying the correct prior is still a highly investigated research topic and several options are discussed as alternatives for isotropic Gaussian priors \cite{priors_bnn}. %The neural network component of UDEs often contains only 10–150 parameters. 
One comparatively simple option is a Gaussian prior with a non-diagonal covariance matrix, allowing for correlation between different parameters \cite{pml2Book}. 
For mechanistic parameters, knowledge about the interpretation of different parameter values often allows for a handcrafted design of prior distributions. %If none is given, a log-uniform distribution with broad enough upper and lower bounds is the standard procedure when exploring several orders of magnitude. 

\begin{figure*}[ht]
    \centering
    \includegraphics[width=0.9\columnwidth]{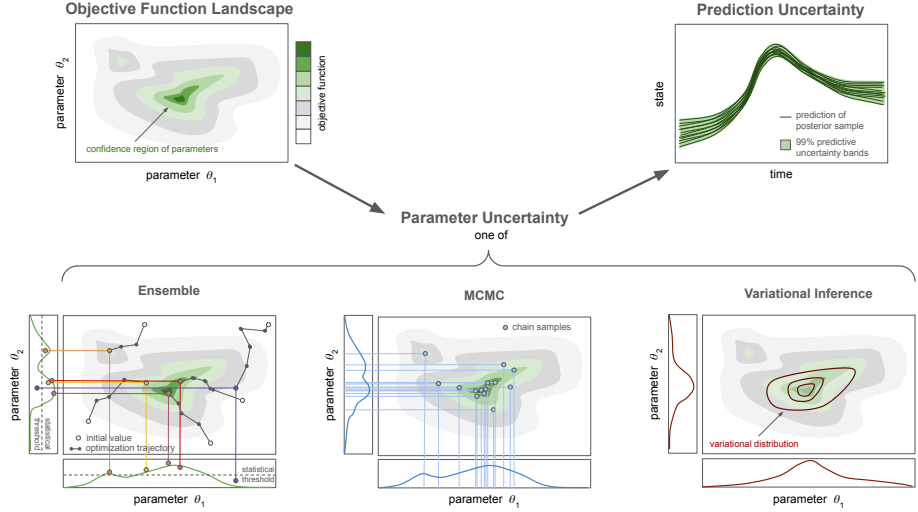}
    \caption{Overview of the presented uncertainty quantification methods. Given an unknown objective function landscape, we estimate the posterior distribution of the parameters using UDE ensembles, MCMC methods or Variational Inference. Based on random draws from the posterior distribution, predictions of the state trajectory are made, yielding lower and upper bounds for a 99\% prediction interval.}
    \label{fig:UQ_methods}
\end{figure*}

For this work, we identified methods that are - from a theoretical point of view - more suitable for the epistemic UQ of UDEs. We consider Profile Likelihood (PL) calculation \cite{kreutz_profile_2013} or asymptotic confidence intervals via the Fisher Information Matrix (FIM) to be less suitable (see \cref{apdx:method_identification} for a discussion) and therefore did not investigate them empirically. However, multistart ensembles, Variational Inference and MCMC-based sampling provide a more promising basis. \cref{fig:UQ_methods} gives a high-level overview of these methods. We provide in-depth descriptions in the following subsections, starting with ensemble-based UQ.

% further ideas for this section: dropout

\subsection{Ensemble-based uncertainty analysis}

Let $m$ be the number of potential models in the ensemble. We assume that each model has the same form, i.e., it shares the same formulation of the differential equation, including the same architecture of the neural network, while its parameter values may differ. Each ensemble member is characterised by its set of parameters $\hat{\bm{\theta}}^i$ for $i \in \{1,...,m\}$, where $\hat{\bm{\theta}}^i$ is obtained by one realisation of a training schedule with random components.
In the Bayesian setting, these estimated parameters can be interpreted as samples from the posterior distribution \cite{pml2Book}. Hence, a sufficiently large number of ensemble members can be utilized for a Monte-Carlo approximation of $p(\bm{\theta}|\mathcal{D})$. 

We propose to combine the approaches from deep learning \cite{lakshminarayanan_simple_2017} and mechanistic dynamical modelling \cite{villaverde_assessment_2023} to define ensemble members that are customized to UDEs. 
After defining a prior distribution, encoding  mechanistic knowledge when possible, we sample $m$ initial values $\Theta_{\text{init}} = \{\bm{\theta}_{\text{init}}^1, ... \bm{\theta}_{\text{init}}^M\}$ from it. Then, $m$ UDEs are trained, each starting with one element of $\Theta_{\text{init}}$. The resulting optima are $\Theta = \{\hat{\bm{\theta}}^1, ..., \hat{\bm{\theta}}^m\}$. During the optimisation procedure, two issues have to be solved: overfitting and non-optimal local minima. To overcome overfitting, we use early stopping (in combination with a L2 regularisation of the neural network parameters). Note that the necessary train-validation split yields a second source of randomness in our ensemble implementation. Like many dynamical approaches, UDEs can face convergence issues and numerical instabilities, which results in estimates with low likelihood. To address the issue of ending the optimization at non-optimal local minima, only a subset of the resulting estimators in $\Theta$ is accepted as ensemble members. We assume that $m$ is large enough, such that the estimated maximum-likelihood estimate over all elements in $\Theta$, $\hat{\bm{\theta}}_{\text{MLE}}$, is approximately equal to the theoretical maximum-likelihood estimate. 
Based on a likelihood-ratio test, we evaluate whether the likelihoods of the other parameters in $\Theta$ significantly differ from  $\hat{\bm{\theta}}_{\text{MLE}}$ (a method commonly used in Systems Biology \cite{kreutz_profile_2013, villaverde_assessment_2023}). Hence, we keep those models whose parameter values are considered to be close enough to a mode of the likelihood distribution. The test statistic is given by
\begin{equation}
    \lambda(\bm{\theta}) = -2 (\mathrm{log}(p(\bm{\hat{y}}|\bm{\theta}_{\text{MLE}}))- \mathrm{log}(p(\bm{\hat{y}}|\bm{\theta})))  
\end{equation}
and evaluated for all $\hat{\bm{\theta}} \in \Theta$.
Asymptotically, the threshold is given by the $\alpha$-quantiles of a $\chi^2$ distribution with $n_f$ degrees of freedom \cite{villaverde_assessment_2023}, where $n_f=1$ can provide a lower bound on the uncertainty. % Asymptotically here means for a sufficiently large number of data points. The quantiles can be calculated by the inverse cumulative density function. 
% Although bad local minima can be an issue for classical neural networks, too, we are - to the best of our knowledge - the first ones to propose an ensemble selection based on the likelihood ratios in the context of UDEs and neural networks. \todo{check once more for NN that this statement is true}

\subsection{Bayesian universal differential equations}

Bayesian models perform Bayesian inference for the parameters of a model based on Eq.~(\ref{equ:Bayes}). Usually, the posterior has no closed-form solution. Instead, one relies on approximate inference algorithms like variational inference or MCMC-based methods \cite{pml2Book}. 

Variational inference approximates the intractable distribution $p(\bm{\theta}|\mathcal{D})$ using a parametric distribution $q(\bm{\theta}|\bm{\psi})$ with distribution parameters $\bm{\psi} \in \mathbb{R}^{n_{\psi}}$ \cite{gawlikowski2022survey}. The Kullback-Leibler divergence $D_{\mathbb{KL}}$ is commonly used as an objective function, describing the discrepancy between the two distributions.
This reduces the inference problem to defining an appropriate variational distribution $q(\bm{\theta}|\bm{\psi})$ and estimating its parameters $\bm{\psi}$. Commonly used options are, for example, a Gaussian distribution $q(\bm{\theta}|\bm{\psi}) = \mathcal{N}(\bm{\theta}|\mu, \Sigma)$, if no parameter bounds are given, or a scaled beta distribution $q(\bm{\theta}|\bm{\psi}) = c \cdot \mathrm{Beta}(\bm{\theta}|a,b)$ if $\bm{\theta} \in (0,c)$.

MCMC methods are algorithms that sample from the posterior distribution, which are then used to create a Monte Carlo approximation of the posterior distribution. Hamiltonian Monte Carlo (HMC) algorithms leverage gradient information to search the parameter space efficiently. HMC algorithms are often used for both dynamical modelling \cite{pypesto} and Bayesian Neural Networks \cite{pml2Book}. Defining efficient sampling methods is an active field of research. It has been shown that the warm-up phase of the sampling algorithm can be reduced when starting the algorithm from optimization endpoints \cite{Ballnus2017-yb}. To mitigate the problem that plain HMC is highly sensitive to parameters of the algorithm, one can use the HMC extension No-U-Turn Sampler (NUTS) \cite{nuts}. At the moment, the NUTS algorithm is state-of-the-art for Bayesian neural networks \cite{sommer2024connecting} and showed the best performance in the context of Neural ODEs \cite{dandekar2022bayesian}. Yet, even with these approaches, the sampler often struggles to explore more than one mode. Parallel tempering algorithms operate with chains on different kinetic energy levels (temperatures). While low-temperature chains explore individual modes, high-temperature chains more easily traverse through the parameter space. By swapping the states of different chains under certain circumstances, parallel tempering algorithms aim to explore multimodal distributions more efficiently \cite{syed_non_reversible_pt_2022}.

% similar methods for dynamical models exist
% For UDEs: basically reuse existing methods

\section{Performance evaluation of methods: insights from synthetic examples}
\label{sec:4}
To assess the performance of the aforementioned UQ methods, we performed experiments on several synthetic problems of increasing complexity. Working with synthetic problems allows us to evaluate the performance of the methods by comparing their results to the data-generating process. 

%\subsection{Reference Uncertainty}
%Synthetic data allows us to evaluate the performance of methods using more information than a real-world scenario could provide. Like in our problem setup, modelers are often faced with sparse and noisy data. The question that we aim to answer with our reference uncertainty is: Would the estimate of epistemic uncertainty look different, if we had chosen more, or a different draw from our noise model? For this, we repeat the ensemble based uncertainty quantification for 30 independently drawn synthetic datasets that follow the same distribution. The reference uncertainty is the combined uncertainty of those 30 realisations. 

\subsection{Model formulation}
We generated synthetic data based on three different equations (SEIR Pulse, SEIR Waves and Quadratic Dynamics) and the two noise models described in \cref{sec:noise_model} (Gaussian and negative Binomial). For each noise model, we investigated two noise parameter settings. \cref{tab:problem_scenarios} in \cref{apdx:problem_overview} gives an overview of the ten considered problem scenarios.
In the main part of the paper, by way of example, we mainly present the results for two out of the three investigated differential equations and hence only define the presented ones here (see \cref{supp:quadratic_dynamics} for the third). We used a common differential equation of epidemiology for the definition of the first two problems:
The SEIR model is a compartmental model to describe the dynamics of infectious diseases in a population. Susceptibles (S) may become Exposed (E) and then become Infected (I) before they Recover (R) from the disease \cite{Tang2020-yu}. We chose the SEIR model as it is a good example of the variety of options one has when encoding prior knowledge into the design of a UDE. The dynamics are governed by
\begin{align}
\label{equ:SEIR}
\begin{split}
\frac{dS}{dt} &= -\beta(t) \frac{S I}{N}, \quad \quad \frac{dE}{dt} = \beta(t) \frac{S I}{N} - \alpha E, \\[2ex]
\frac{dI}{dt} &= \alpha E - \gamma I, \quad \quad \: \frac{dR}{dt} = \gamma I,
\end{split}
\end{align}
where $\beta$ is the transmission rate, $\alpha$ the transition rate, $\gamma$ the recovery rate,  and $N = S + E + I + R$ the population size.
We create synthetic data with two different settings of the time-varying transmission rate $\beta(t)$ (see \cref{sup-fig:overview_beta} for a visualisation).

For the \textbf{SEIR Pulse} scenario, we defined the underlying $\beta$ as 
\begin{equation}
    \beta(t) = \begin{cases}
    0.5\qquad  \ \mathrm{if }\ 15<t<30, \\
    0.05\qquad  \mathrm{else.}
    \end{cases}
\end{equation}
This can represent, for instance, political and time-restricted interventions. 

In the \textbf{SEIR Waves} scenario, instead of using a step-wise function for the transmission rate, we used a periodic function with decreasing frequency to create synthetic data:
\begin{equation}
    \beta(t) = \mathrm{cos} \left( (-1 + \sqrt{1+4t}) \cdot 1.5 + 0.25 \cdot \pi \right) \cdot 0.3 + 0.4.
\end{equation}
This function demonstrates an exemplary complex change of the transmission rate with several waves that may occur due to new virus variants, and changing behaviour of the population or vaccination levels. 

For the synthetic data generation, we assumed that only measurements of the states $I$ and $R$ can be observed. For each problem setting and noise scenario, 30 measurement data points uniformly spaced in time are created based on noise-injection after solving the differential equations using fixed mechanistic parameters (see \cref{tab:ground_truth_parameters_overview}). In the scenario of Gaussian noise, we modelled the fraction of infections with $\bm{x}_0 = (S(0), E(0), I(0), R(0)) = (0.995, 0.004, 0.001, 0.0)$. In the case of a Negative Binomial noise model, we described absolute numbers with $\bm{x}_0 = (995, 4, 1, 0)$. 
We analysed UDEs where it is assumed that the mechanistic formulation of the differential equation is known, but the time-varying transmission rate $\beta$ is unknown. Therefore, a neural network was used to model $\mathrm{log}(\beta)$, where we use a log-parameterization to ensure positivity and easily capture fast-changing dynamics. The precise values of the other mechanistic parameters $\alpha$ and $\gamma$ were assumed to be unknown, but realistic bounds were enforced using a transformation that is based on the $\mathrm{tanh}$ function (see \cref{apdx: implementation details}).

\subsection{Results}
\label{sec:implementation and results}

We start the investigation of UQ by explaining and investigating each of the three discussed methods (ensembles, MCMC, and Variational Inference) individually. Afterwards, we compare their performance and mode exploration. Implementation details are provided in \cref{apdx: implementation details} and additional figures in \cref{apdx:additional_figures}. 

\subsubsection{Ensemble-based uncertainty}

\cref{fig:seir_overview,fig:seir_overview_parameters,fig:seir_beta_overview} provide an overview of the results for the SEIR Waves problem with a Gaussian noise model for $\sigma=0.01$ using the presented ensemble-based UQ method. 
In general, the UQ worked reasonably well: Observed states show a smaller predictive uncertainty than the unobserved states. 
The uncertainty bands for parameters for which we had more informative data ($ \gamma $ can be derived from $I$ and $R$) are smaller than for those for which we did not have such detailed information ($\alpha $). We observe in \cref{fig:seir_overview_parameters} that the data generating value of the standard deviation $\sigma$ lies within the estimated posterior distribution and its mode. Yet, the long tail of the distribution indicates that, on rare occasions, the aleatoric uncertainty was overestimated.
While the underlying dynamics of the unobserved states $S$ and $E$ could be recovered, this was not the case for $\beta$ (see \cref{fig:seir_beta_overview}). A broad band of trajectories of $\beta$ yields reasonable values for the observed states $I$ and $R$. Since $\beta$ influences the dynamics only in scenarios where $I\cdot S >> 0$, we would only expect reasonable estimates in this regime. Ensemble members with smaller negative log-likelihood values tend to show dynamics more closely related to the dynamics of the data generation process for these time points. Outside the estimatable region, the neural network tends to output more constant trajectories, which may be due to the implemented L2 regularization.

In this context, it should be noted that we also tried out the estimation of a constant $\beta$: Instead of using a neural network, we treated $\beta$, similar to $\gamma$ and $\alpha$, as a constant parameter. A constant $\beta$ was not able to describe the data reasonably well (see  \cref{fig:seir_constant_beta}). 
 \cref{fig:seir_pulse_noise_model} displays the trajectories of the state variables for the three noise settings in the SEIR Pulse scenario. We observe that, as expected, the ensemble-based method yields larger prediction uncertainty bounds with increasing aleatoric noise. Small fluctuations in the trajectory cannot be captured easily within a setting of negative binomial noise, as is indicated by the ensemble mean trajectory for $I$. %This is not only displayed in the trajectories of the states but can also be seen in a broader distribution of the noise parameters. 
In the SEIR problem scenarios, the role of the neural network is well isolated from the other dynamical components. In the Quadratics Dynamics scenario, this is different, because the neural network can in principle completely replace known mechanistic dynamics and describe all the dynamics. Hence, the mechanistic part is only a soft constraint on the form of the whole dynamics. A consequence of this is visualized in \cref{fig:quadratic_dynamics}: The predictions of some ensemble members quickly deviate from the reference dynamics outside the data domain.

One difficulty of the ensemble-based method is the arbitrarity of choosing a reasonable threshold. As visualized in \cref{fig:seir_overview}, subselecting a fraction of the best performing models (which is equivalent to a different significance level for the $\chi^2$-test) can result in widely different confidence bands. An exemplary waterfall plot displaying the selection of the ensemble members based on likelihood values is provided in \cref{fig:waterfall_plot1} and shows that there is no clear convergence to a minimum objective function value for UDEs. 
A major advantage of the ensemble-based uncertainty quantification method is its flexible parallelizability: Every candidate member of the ensemble can be trained independently of one another. 
For 10~000 candidate ensemble members, the training took between 4–12 hours using 20 CPU cores.

\subsubsection{Bayesian UDEs}

\begin{figure*}[t]
\vskip 0.2in
\begin{center}
\centerline{\includegraphics[width=1\textwidth]{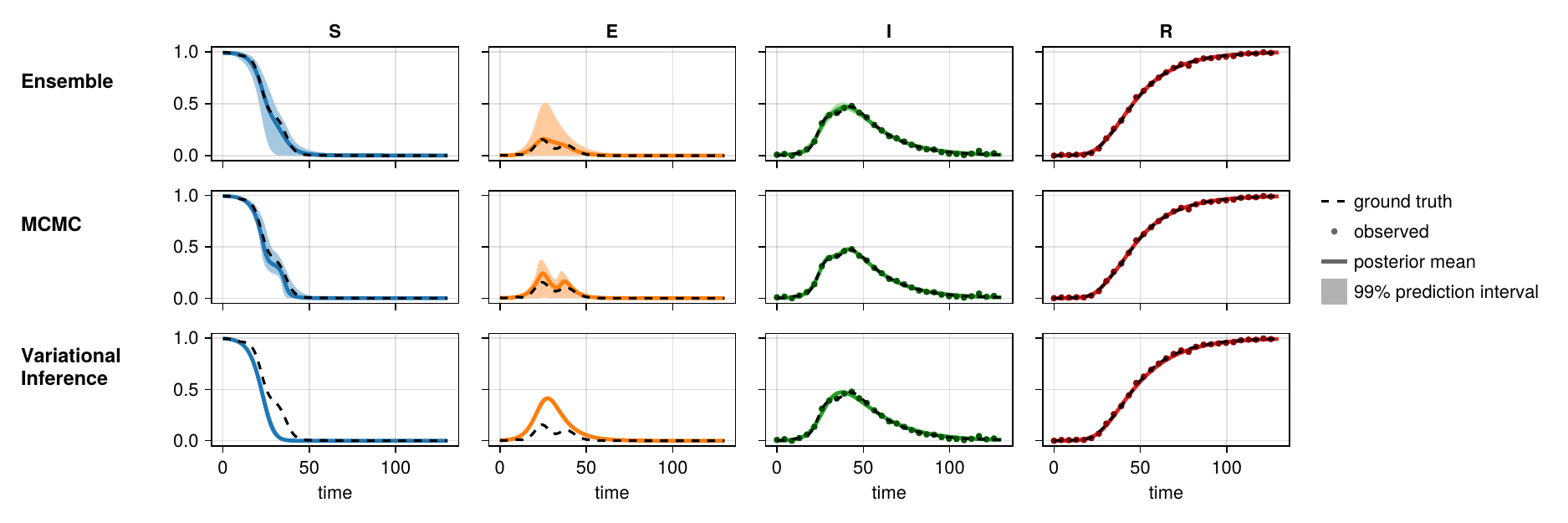}}
\caption{Comparison of the UQ methods on the SEIR Waves problem (Gaussian noise, $\sigma=0.01$).}
\label{fig:method_comparison}
\end{center}
\vskip -0.2in
\end{figure*}

We implemented Bayesian UDEs using a No-U-Turn (NUTS) and parallel tempering sampling algorithm to compare different and potentially suitable algorithms for UDEs. 
% At the moment, the NUTS algorithm is state-of-the-art for Bayesian neural networks \cite{sommer2024connecting} and showed the best performance in the context of Neural ODEs \cite{dandekar2022bayesian}. 
When sampling, the biggest issue of overparametrized models is the exploration of multiple modes. Neural networks specifically tend to have by construction various symmetries in the loss landscape \cite{wimmer_2023_nn_symmetries}, resulting in the possibility that no additional predictive information is added even if multiple modes are explored. We systematically experimented with different numbers of chains and samples. However, similar to what is observed with neural networks in classical supervised learning tasks \cite{sommer2024connecting}, MCMC chains do neither mix well nor properly converge in the context of UDEs. Based on a simple clustering analysis on the samples, we found out that a NUTS algorithm with 7 chains and 100~000 samples tends to explore slightly fewer modes than a parallel tempering algorithm that only returns one chain in the parameter space and returns only 10~000 samples. Furthermore, the distributions of the mechanistic parameters are smoother and narrower in the parallel tempering case. 
Starting the sampling process from optimization endpoints is a long-standing rule in classical dynamical modelling \cite{hass_2019_benchmark_problems} and has shown first advantages in the field of deep learning \cite{sommer2024connecting}. Since UDEs are build upon these two concepts, it is not surprising that we have had good experiences using this warm-start method in the context of UDEs, too.
\cref{fig:method_comparison} visualizes the results of the parallel tempering algorithm, indicating reasonable fits. While very precise and narrow distributions are the result of this method for the noise parameter and $\gamma$, we observed comparatively broad bands for $\alpha$ (see \cref{fig:constant_parameters_MCMC}). For approximately 10 000 samples, the algorithm required 5-7 days of computation on 20 CPU cores. 

\subsubsection{Variational Inference}
The most important and limiting hyperparameter of Variational Inference is the choice of variational distributions used to approximate the posterior distribution. While having many drawbacks, the mean-field approximation with multivariate normal base distributions is considered one of the standard methods to perform Variational Inference and returns plausible results of the predictive uncertainty for many problems \cite{ABDAR2021243}. \cref{fig:method_comparison} visualizes the results of the Variational Inference method applied to the SEIR problem. The observed states can be described reasonably well. Yet, the uncertainty bands of observed and unobserved states do not differ concerning their width, indicating that only one mode was captured. Since the reference line of the unobserved states is not covered, we expect that there exist at least two modes in the loss landscape that can explain the observed data. 

The limited flexibility of the variational distribution tends to prevent overfitting. Yet, in the context of dynamical modelling and specifically UDEs, Variational Inference with a mean-field approximation is a too limited approach to capture the underlying uncertainties. %The multivariate Gaussian explored only one mode, which is especially problematic in describing the uncertainty of the unobserved states or in scenarios of multi-modal parameter distributions. 
The training process is in comparison to ensemble or sampling-based methods much faster (<24 h on one CPU core for the considered models). 

\subsubsection{Comparison of methods}
The three presented methods tackle the problem of UQ from different angles. This results in construction differences concerning the data usage and incorporation of prior knowledge, as well as expected differences concerning the number of covered modes. Since the ensemble-based method is based on the optimization of an over-parametrized model, a train-validation split is necessary to implement early stopping and avoid overfitting on the training data. The combined dataset is only used for subselecting from ensemble candidate models. For MCMC-based methods and Variational Inference, the whole dataset is used in every step of the algorithms. 
While the incorporation of prior knowledge in the dynamic equations, noise model, and observable mapping is independent of the UQ method, assumptions about parameter values are treated differently. For the ensemble-based method, the parameter prior influences the start points of the optimization process. Afterwards, we only encode upper and lower bounds that restrict the parameter update steps. For both MCMC and Variational Inference, the prior distributions influence the posterior throughout the algorithms. 

\cref{fig:method_comparison} compares the three methods on the SEIR problem. As discussed in the previous subsection, the implemented Variational Inference method is not capable of capturing the underlying hidden dynamics. The uncertainty of the MCMC method is smaller than the uncertainty of the ensemble-based method. This can have two main reasons: Either the MCMC method does not cover as many modes or the threshold for selecting the ensemble members was chosen to be suboptimal. While we have distributional guarantees for the threshold for large sample sizes, this is not necessarily the case for limited sample sizes \cite{kreutz_2023_lr_finite_sample} and, while it is essential to define a threshold, there is still no optimal method available. 

\begin{figure*}[t]
\vskip 0.2in
\begin{center}
\centerline{\includegraphics[width=1\textwidth]{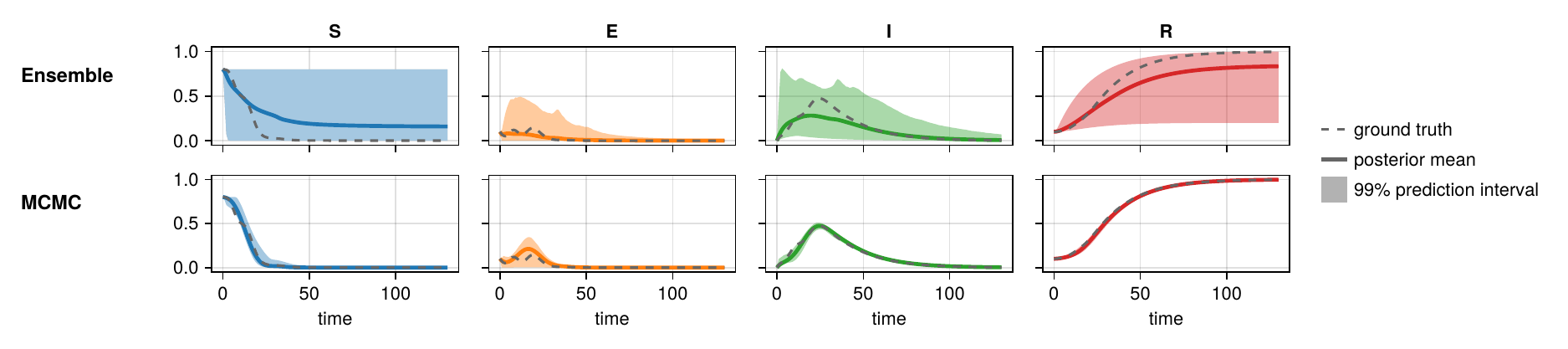}}
\caption{Comparison of prediction performance using the same setting as in \cref{fig:method_comparison} but with a new initial condition $\bm{x}_0 = (0.8, 0.1, 0.0, 0.1)$.}
\label{fig:IC_method_comparison}
\end{center}
\vskip -0.2in
\end{figure*}

To investigate the different results of the ensemble and MCMC-based methods more thoroughly, we looked at the performance of the method when applying it to a new initial condition. If all parameters and $\beta(t)$ are learned correctly, the method should be capable of providing good fits to this new reference trajectory. \cref{fig:IC_method_comparison} shows that only the MCMC-based method is capable of this. To investigate the difference in the posterior parameter samples of the two methods, we applied a UMAP analysis. Here, each parameter dimension corresponds to one feature of the UMAP input dataset. The chosen threshold of the ensemble-based method leads to parameter values that cluster together, while the parallel-tempering algorithm yields several separate clusters and explores more modes (see Appendix \ref{apdx:UMAP}).

\section{Conclusions and future perspectives}
UQ for UDEs yields a unique challenge: Overparametrized models with a subset of interpretable parameters that describe both observed and unobserved state trajectories. Both ensemble and MCMC-based methods show better UQ performance than Variational Inference in our conducted experiments. Yet, the definition of a suitable threshold for ensemble-based methods needs to be further investigated before applying the method to new initial conditions. Apart from this, we propose three future research directions. Firstly, the development of hybrid UQ approaches, where one tries to combine the advantages of resource-efficient ensemble methods with the precision of MCMC methods. Secondly, building upon the work of \cite{sommer2024connecting}, symmetry removal in the objective landscape prior to UQ might improve computational efficiency and seems, in the context of the commonly used small neural networks for UDEs, more feasible than for larger neural networks. Lastly, investigating model form uncertainty may provide interesting insights into the phenomenon of the absorption of mechanistic terms by the neural network.

%\begin{itemize}
%    \item Problem of the definition of a reference: the optimization method influences the distribution, so if we use an optimization method for the reference that is more similar to on of the UQ methods it is not a fair comparison. There is no ground-truth epistemic uncertainty. 
%\end{itemize}

%Potential drawbacks of Ensemble-based method:
%\begin{itemize}
%    \item ensemble size is not pre-determined \cite{villaverde_assessment_2023}
%    \item If the optimal parameter value is located close to the corners of a bounded parameter space, simply by geometry, we're less likely to find the correct parameter values.
%\end{itemize}

% \section*{Software}
%The code used for the experiments discussed above is available under \todo{add github link}. 

\section*{Acknowledgements}
This work was supported by the Deutsche Forschungsgemeinschaft (DFG, German Research Foundation) under Germany’s Excellence Strategy (EXC 2047—390685813, EXC 2151—390873048), by the German Federal Ministry of Education and Research (BMBF) (INSIDe - grant number 031L0297A), and by the University of Bonn (via the Schlegel Professorship of J.H.). W.W. received funding from the Flemish Government under the ``Onderzoeksprogramma Artifici\"le Intelligentie (AI) Vlaanderen" Programme.

\bibliography{submitted_paper}
\bibliographystyle{icml2024}

%%%%%%%%%%%%%%%%%%%%%%%%%%%%%%%%%%%%%%%%%%%%%%%%%%%%%%%%%%%%%%%%%%%%%%%%%%%%%%%
%%%%%%%%%%%%%%%%%%%%%%%%%%%%%%%%%%%%%%%%%%%%%%%%%%%%%%%%%%%%%%%%%%%%%%%%%%%%%%%
% APPENDIX
%%%%%%%%%%%%%%%%%%%%%%%%%%%%%%%%%%%%%%%%%%%%%%%%%%%%%%%%%%%%%%%%%%%%%%%%%%%%%%%
%%%%%%%%%%%%%%%%%%%%%%%%%%%%%%%%%%%%%%%%%%%%%%%%%%%%%%%%%%%%%%%%%%%%%%%%%%%%%%%
\newpage

\appendix
\beginsupplement
\onecolumn
\section{Discussion of additional UQ methods} 
\label{apdx:method_identification}
Some options that are commonly used for UQ in mechanistic modelling, like those based on the Fisher Information Matrix (FIM) and Profile Likelihoods (PLs), are not applicable to UDEs. Others, like ensemble-based and fully Bayesian settings, have more potential and hence, have been investigated thoroughly in the main part of this paper.

Asymptotic confidence intervals via the FIM \cite{kreutz_profile_2013} are a method commonly used in the context of dynamical modelling to evaluate the covariance matrix of a maximum likelihood estimate. The simplicity of this idea is convincing in the setting of identifiable parameters. Yet, it is not suitable once $ p(\bm{\theta} |\mathcal{D} )$ has a plateau of optimal values. The inverse of the hessian and hence, the FIM is not well defined in these settings. UDEs share the same issues as neural networks: Overparametrization of parameters and plateaus in the loss space are very likely. 

PL \cite{JACQUEZ1985201} evaluate the shape of $p(\bm{\theta}|\mathcal{D})$ by fixing one parameter $\theta_i$ to a value $z$ and re-evaluating the best maximum-likelihood estimate feasible for this fixed value, i.e. 
\begin{align}
    \mathrm{PL}(z) = \underset{\bm{\theta} \in \{\bm{\theta} | \theta_i=z \} }{\max} p(\mathcal{D}|\bm{\theta}).
\end{align}
This procedure is done multiple times, scanning over a series of fixed values $z$. Profile likelihoods are a comparatively fast method to evaluate plausible values in one dimension of $\bm{\theta}$. For predictive uncertainty, all dimensions of $\bm{\theta}$ have to be evaluated, which is still comparatively fast if the parameters are not correlated. Yet, this becomes infeasible for higher dimensions of $\bm{\theta}$ that are highly correlated. Hence, while profile likelihoods can be used to get a deeper understanding of the uncertainty of the mechanistic parameters of the UDE, they are not suitable for the evaluation of prediction uncertainty of UDEs.

\section{Problem overview}
\label{apdx:problem_overview}

\begin{figure}[ht]
\vskip 0.2in
\begin{center}
\centerline{\includegraphics[width=0.7\textwidth]{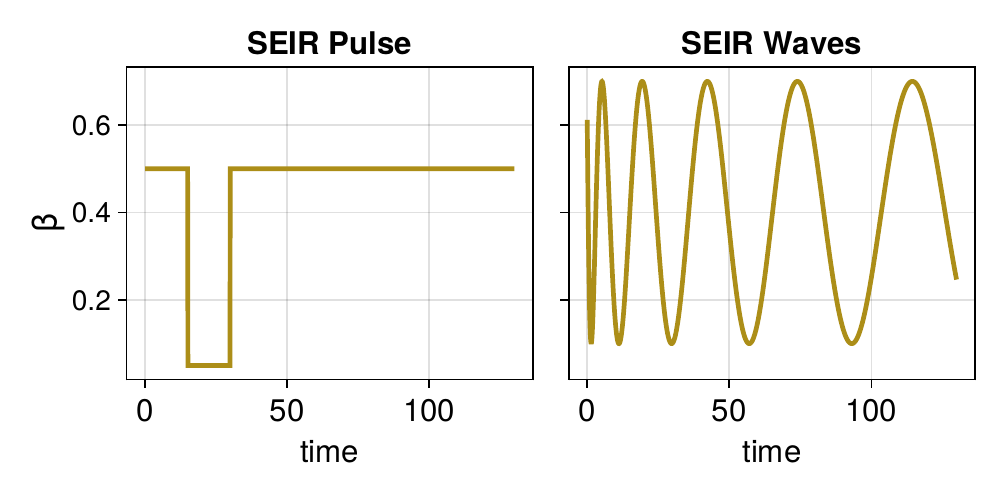}}
\caption{Visualization of the values of $\beta$ for the data generation process of the SEIR Pulse and SEIR Waves problem scenarios.}
\label{sup-fig:overview_beta}
\end{center}
\vskip -0.2in
\end{figure}

In the following, we list a few tables that provide an overview over the different problem scenarios, its initial conditions and parameter values. Furthermore, we provide a visualisation of the values of $\beta$ for the SEIR Waves and SEIR Pulse settings in \cref{sup-fig:overview_beta}.

\begin{table*}[h]
\caption{Overview of the synthetic problem scenarios considered, including information about the time span, number of time points $n_t$ and parameters $n_{\bm{\theta}}.$}
\label{tab:problem_scenarios}
\vskip 0.15in
\begin{center}
\begin{small}
\begin{sc}
\begin{tabular}{lccccc}
\toprule
Dynamic model & Noise model & Noise parameter & time span & $n_t$ & $n_{\bm{\theta}}$ \\
\midrule
\multirow{2}{*}{Quadratic Dynamics} & \multirow{2}{*}{Gaussian} & $\sigma=0.01$ & \multirow{2}{*}{$(0.0,10.0)$} & \multirow{2}{*}{12} & \multirow{2}{*}{22} \\
&  & $\sigma=0.05$ & & & \\
\midrule
\multirow{4}{*}{SEIR Waves} & \multirow{2}{*}{Gaussian} & $\sigma=0.01$ & \multirow{4}{*}{$(0.0,130.0)$} & \multirow{4}{*}{30} & \multirow{4}{*}{64} \\
&  & $\sigma=0.05$ & & \\
& \multirow{2}{*}{Negative Binomial} & $d=1.2$ & & \\
&  & $d=2.2$ & & \\
\midrule
\multirow{4}{*}{SEIR Pulse} & \multirow{2}{*}{Gaussian} & $\sigma=0.01$ & \multirow{4}{*}{$(0.0,130.0)$} & \multirow{4}{*}{30} & \multirow{4}{*}{64} \\
&  & $\sigma=0.03$ & & \\
& \multirow{2}{*}{Negative Binomial} & $d=1.2$ & & \\
&  & $d=2.2$ & & \\
\bottomrule
\end{tabular}
\end{sc}
\end{small}
\end{center}
\vskip -0.1in
\end{table*}

\begin{table*}[h]
\caption{Overview of the initial conditions for the synthetic problem scenarios considered.}
\label{tab:IC_overview}
\vskip 0.15in
\begin{center}
\begin{small}
\begin{sc}
\begin{tabular}{lccc}
\toprule
Dynamic model & Noise model & Initial condition \\
\midrule
Quadratic Dynamics & Gaussian  & $(0.1)$ \\
\midrule
\multirow{2}{*}{SEIR Waves} & Gaussian  & $(0.995, 0.004, 0.001, 0.0)$ \\
& {Negative Binomial} & $(995.0, 4.0, 1.0, 0.0)$ \\
\midrule
\multirow{2}{*}{SEIR Pulse} & Gaussian  & $(0.995, 0.004, 0.001, 0.0)$ \\
& {Negative Binomial} & $(995.0, 4.0, 1.0, 0.0)$ \\
\bottomrule
\end{tabular}
\end{sc}
\end{small}
\end{center}
\vskip -0.1in
\end{table*}

\begin{table*}[h]
\caption{Overview of the mechanistic parameters of the differential equations for the synthetic problem scenarios considered.}
\label{tab:ground_truth_parameters_overview}
\vskip 0.15in
\begin{center}
\begin{small}
\begin{sc}
\begin{tabular}{lcccc}
\toprule
Dynamic model & $\alpha$ & $\beta$ & $\gamma$\\
\midrule
Quadratic Dynamics & $1.0$ & $2.0$ & -\\
SEIR Waves & $0.9$ & - & $0.1$ \\
SEIR Pulse & $0.33$ & - & $0.05$ \\
\bottomrule
\end{tabular}
\end{sc}
\end{small}
\end{center}
\vskip -0.1in
\end{table*}

\clearpage

\section{Quadratic dynamics}
\label{supp:quadratic_dynamics}
% \subsection{Problem definition}
This problem is a comparatively small problem with only two non-identifiable mechanistic parameters. While finding local optima in the loss space should be feasible, the identification of distributions of non-identifiable parameters is more complex.

To generate synthetic data for the quadratic dynamic problem, we simulate the differential equation
\begin{align}
\label{equ:quadratic_dynamics}
\begin{split}
    \frac{dx}{dt} &= \alpha x - \beta x^2, \\
    x(0) &= 0.1
\end{split}
\end{align}
with $t \in (0, 10)$, $\alpha = 1$ and $\beta=2$. We assume, that the observable mapping is the identity, i.e. $x$ is directly observed. Noise is added to the observable $x$ according to Table \ref{tab:problem_scenarios}. We assume that only one of the components of the mechanistic terms in the differential equation is known. Hence, the UDE is defined as

\begin{align}
\label{equ:quadratic_dynamics_ude}
\begin{split}
    \frac{d\hat{x}}{dt} &= \hat{\alpha} \hat{x} - f_{\text{net}}(\hat{x};\bm{\theta}_{\text{net}}), \\
    x(0) &= 0.1
\end{split}
\end{align}

where $f_{\text{net}}$ is a fully connected neural network with parameters $\bm{\theta}_{\text{net}}$. To ensure the positivity of $\alpha$ we parametrize $\alpha$ as $\mathrm{log}(\alpha)$. Both the initial value of $\mathrm{log}(\alpha)$ and that of the noise parameter $\sigma$ are sampled from a log-uniform distribution with a minimal value of 0.1 and maximum value of 10.0. 

\section{Implementation details}
\label{apdx: implementation details}
All experiments were conducted on the Unicorn cluster (CPU cores: 2x AMD EPYC 7F72; 3.2 GHz, 1 TB RAM) at the university of Bonn.

Due to its variety of available solvers and automatic differentiation support, we implemented all experiments in Julia \cite{Julia-2017}. The ensemble-based parameter estimation was conducted using packages introduced with SciML \cite{rackauckasUniversalDifferentialEquations2021}. For Variational Inference and most of the MCMC sampling algorithms, we used Turing \cite{ge2018t}. For parallel tempering, the package Pigeons \cite{surjanovic2023pigeons} provided sampling algorithms with an interface to Turing models. A full list of packages is provided in the environment's Manifest and Project files.

As is commonly done in the context of dynamical modelling \cite{hass_2019_benchmark_problems}, we transformed the mechanistic parameters for estimation. The standard deviation was optimized in log scale. Furthermore, we implemented a tanh-based transformation for the other mechanistic parameters to ensure consistency with parametric bounds independent of the optimization algorithm (see Appendix \ref{apdx: tanh} for details). 

The neural network architecture was the same throughout all reported experiments. Specifically, after a short initial hyperparameter search, we used a feed-forward neural network with 2 hidden layers with 6 neurons each and tanh activation functions for all layers apart from the output layer. 

Synthetic data was created according to the problem description, noise model and ground-truth parameters provided in the sections above. All UQ methods used the same data per problem scenario.

\subsection{Ensemble-based UQ}
\label{apdx:details_ensemble}
The neural networks' initial parameter values are sampled according to the default setting (Glorot uniform \cite{glorot_uniform} for weights, zero for biases) with one exception: We observed a more stable training process with fewer numerical instabilities during the solving process of the dynamic equation when the initial parameters of the neural network were initialized to values equal to zero. The mechanistic parameters were sampled according to the prior distributions specified in \cref{apdx:prior_distributions}.
Similar to \cite{rackauckasUniversalDifferentialEquations2021}, optimization was realised using the optimization algorithms ADAM (for the first 4000 epochs) and then BFGS (up to 1000 epochs). To avoid overfitting, we introduced a small L2 regularization on the neural network parameters (with penalty factor $10^{-5}$) and retrospectively stored those parameters per model training that minimized the negative log-likelihood on the respective validation set. The train-validation split was implemented using one individualized random seed per potential ensemble member. The selection of ensemble members from the candidate models was conducted using a significance level of 0.05. 

\subsection{Tanh-based parameter transformation}
\label{apdx: tanh}
While box-constraints are available for many optimization algorithms in Julia, this is not the case for the BFGS algorithm. Ensuring that parameters stay within physically plausible bounds is, however, often necessary to define a solvable differential equation. Furthermore, encoding more prior knowledge can narrow down the hypothesis space of the model and hence, make the exploration of the loss landscape more feasible. 

BFGS is a standard optimization algorithm for UDEs \cite{rackauckasUniversalDifferentialEquations2021}. For purely mechanistic dynamical models, primarily other optimizers that with customized box-constraint implementations are used \cite{pypesto}. We use a tanh-based transformation of the parameters that allows enforcing box-constraints independent of the optimization algorithm:  

Let $\theta^p_i$ be the parametrized version of a mechanistic parameter $\theta_i$. By setting $\theta_i = a \cdot \tanh{(\theta^p_i - c)} + b$ for suitable $a, b, c \in \mathbb{R}$, we can ensure that for any $\theta^p_i \in \mathbb{R}$, $\theta_i$ stays within given bounds. The constant $c$ allows for symmetry around $\theta^p_i=0$.

For the SEIR based problems, the latent period (inverse of $\alpha$) could reasonably be anywhere from an hour (e.g. for certain foodborne illnesses) to several years (e.g. certain malaria cases), hence we assume $\alpha \in (0,24)$ to be known. Similarly, we assume that a person stays infectious for at least one day, i.e. $\gamma \in (0,1)$ and that $\beta(t) \in (0,3)$. As described in \cref{supp:quadratic_dynamics}, no tanh bounds were used for the quadradic dynamics problem.

\subsection{Prior definition for the mechanistic and neural network parameters}
\label{apdx:prior_distributions}
Table \ref{tab:parameter prior} gives an overview of the prior definition of the mechanistic parameters for the different problem scenarios. For Variational Inference and MCMC based sampling, the neural network parameters' prior was defined as $\mathcal{N}(\bm{0}, 3 \cdot I)$.

\begin{table*}[h]
\caption{Overview of the priors for the mechanistic parameters of the considered problems. Note that the noise parameters $\sigma$ and $d$ are only defined in a Gaussian and negative Binomial noise setting, respectively. The mentioned Normal distribution is defined by its mean and standard deviation, the (Log-)uniform distribution by its lower and upper bounds and the Beta distribution by its two shape distributions.}
\label{tab:parameter prior}
\vskip 0.15in
\begin{center}
\begin{small}
\begin{sc}
\begin{tabular}{lcccc}
\toprule
\multirow{2}{*}{Problem} & \multirow{2}{*}{Parameter} & Prior on transformed  & \multirow{2}{*}{Prior} \\
 & & parameter space &  \\
\midrule
Quadratic Dynamics & $\alpha$ & True & $\text{LogUniform}(0.1,10)$ \\
Quadratic Dynamics & $\sigma^2$ & True & Uniform(-10,10) \\
\midrule
SEIR Waves/Pulse & $\alpha$ & True & Normal(0,1) \\
SEIR Waves/Pulse & $\gamma$ & True & Normal(0,1) \\
SEIR Waves/Pulse & $\sigma^2$ & True &  Uniform(-10,10) \\
SEIR Waves/Pulse & $d$ & True & Beta(2,2) \\
\bottomrule
\end{tabular}
\end{sc}
\end{small}
\end{center}
\vskip -0.1in
\end{table*}

\newpage
\section{Additional figures}
\label{apdx:additional_figures}
\counterwithin*{figure}{section}
\renewcommand{\thefigure}{S\arabic{figure}}%
\setcounter{figure}{1}
\subsection{Additional figures for UDE ensembles}
\begin{figure}[ht]
\vskip 0.2in
\begin{center}
\centerline{\includegraphics[width=1\textwidth]{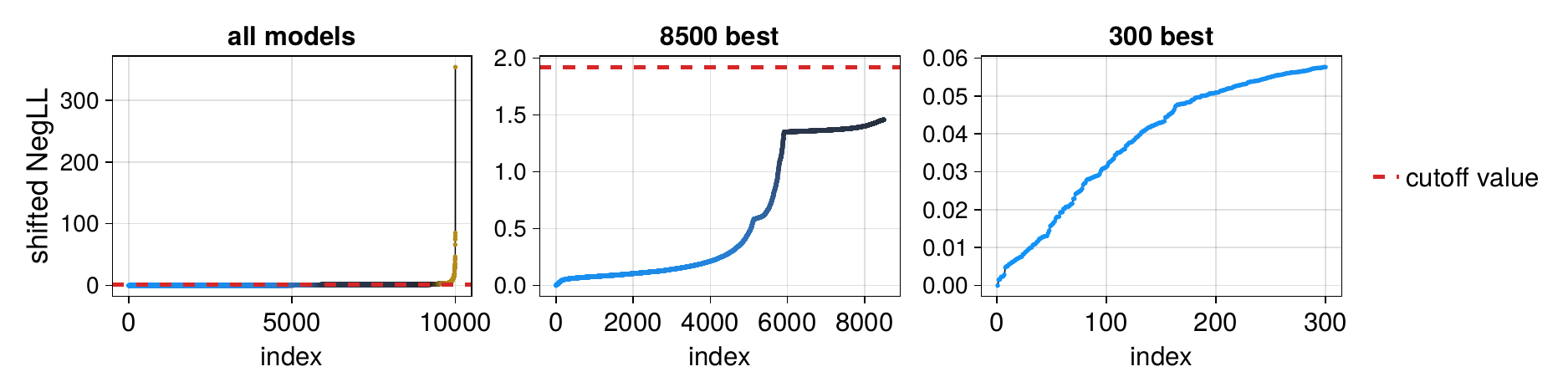}}
\caption{Waterfall plot for the SEIR Pulse problem with Gaussian noise (0.01) for all models (left), the best 8500 models (middle) and the best 300 models (right) according to the negative log likelihood values (NegLL). The y-values are shifted by the minimal value obtained. The cut-off value clearly discards failed model trainings. Unlike many mechanistic systems, UDEs tend to not converge to a global minimum when using multistart optimization.}
\label{fig:waterfall_plot1}
\end{center}
\vskip -0.2in
\end{figure}

\begin{figure}[ht]
\vskip 0.2in
\begin{center}
\centerline{\includegraphics[width=1\textwidth]{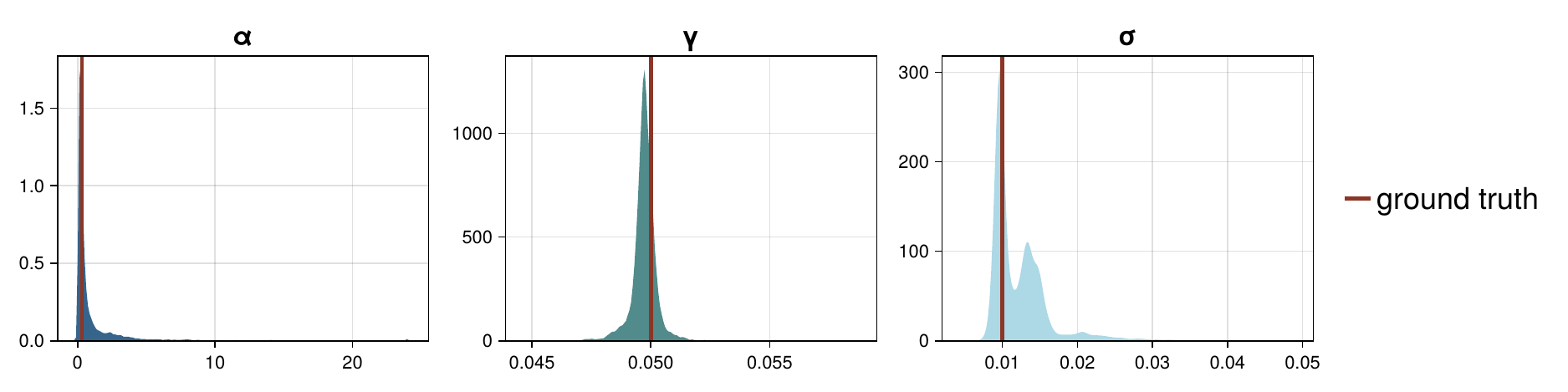}}
\caption{Visualization of the ensemble-based uncertainty quantification for the constant mechanistic parameters (of both the dynamic and noise model) of the SEIR Waves scenario with Gaussian noise (0.01). The red line indicates what parameter value was used to generate synthetic data.}
\label{fig:seir_overview_parameters}
\end{center}
\vskip -0.2in
\end{figure}

\begin{figure}[ht]
\vskip 0.2in
\begin{center}
\centerline{\includegraphics[width=0.9\textwidth]{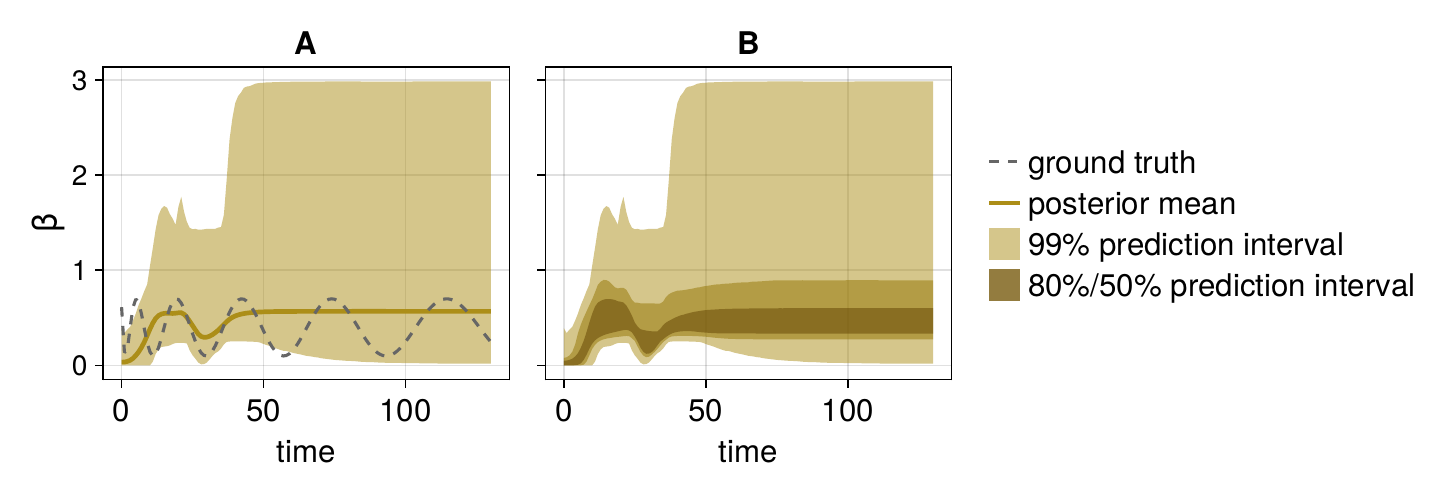}}
\caption{Visualization of the ensemble-based uncertainty of the neural network prediction of $\beta$ for the SEIR Waves scenario with Gaussian noise (0.01).}
\label{fig:seir_beta_overview}
\end{center}
\vskip -0.2in
\end{figure}

\begin{figure*}[ht]
\vskip 0.2in
\begin{center}
\centerline{\includegraphics[width=\textwidth]{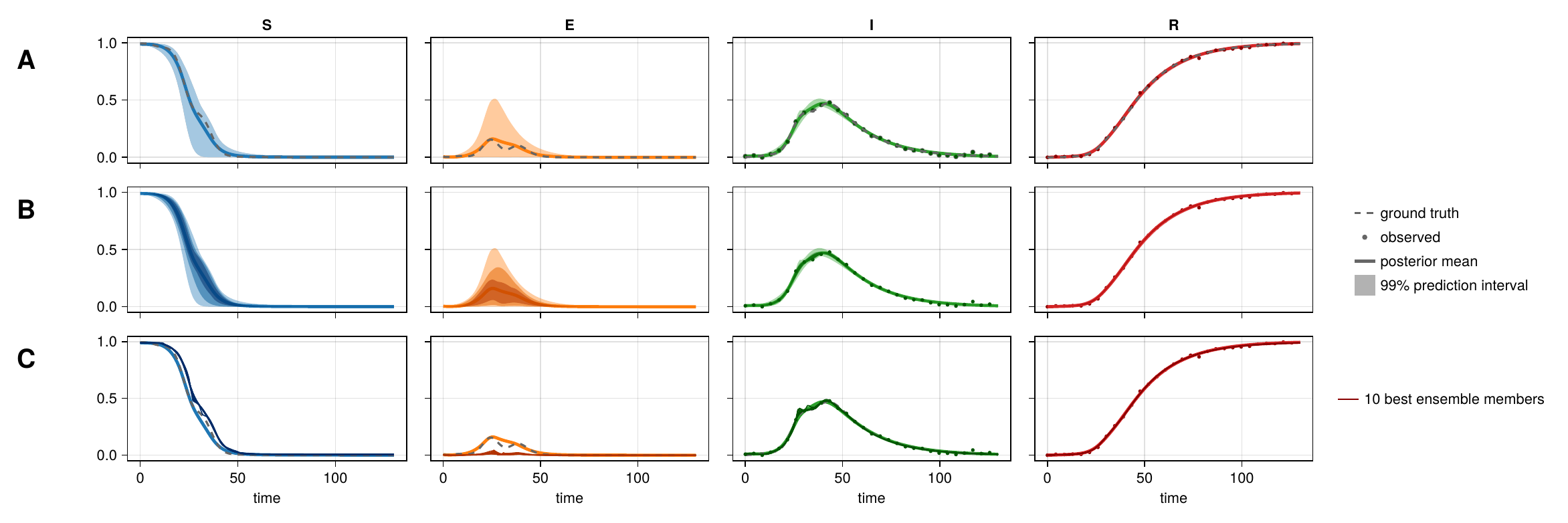}}
\caption{Visualization of the ensemble-based epistemic predictive uncertainty estimation of the SEIR Waves scenario with Gaussian noise (0.01). A: Area covered by all ensemble members per state in comparison to the data generating dynamics (ground truth). B: Area coverages for 99\%, 80\% or 50\% of the best ensemble members. C: Visualization of the trajectories of the 10 best ensemble members (according to the negative Log Likelihood). }
\label{fig:seir_overview}
\end{center}
\vskip -0.2in
\end{figure*}

\begin{figure}[ht]
\vskip 0.2in
\begin{center}
\centerline{\includegraphics[width=1\textwidth]{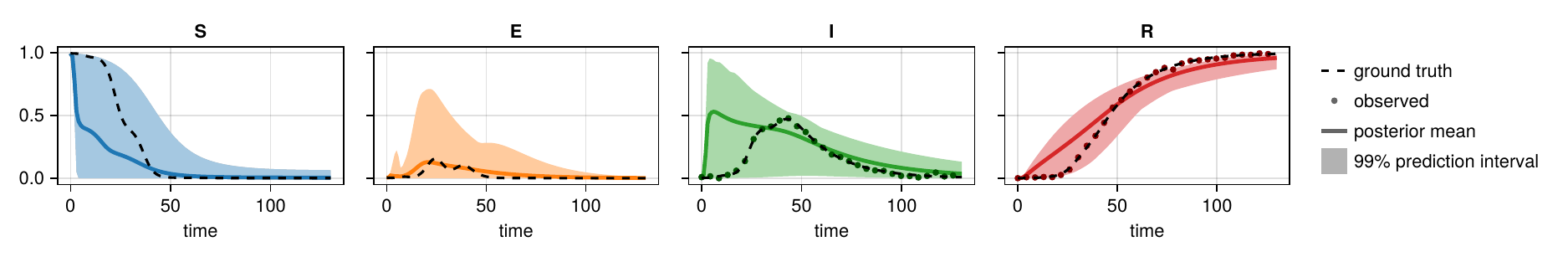}}
\caption{Visualization of the ensemble-based parameter uncertainty estimation assuming a constant $\beta$ instead of the proposed neural network for the SEIR Waves scenario with Gaussian noise (0.01). For observations of the state R lie not within the prediction bands.}
\label{fig:seir_constant_beta}
\end{center}
\vskip -0.2in
\end{figure}

\begin{figure*}[ht]
\vskip 0.2in
\begin{center}
\centerline{\includegraphics[width=1\textwidth]{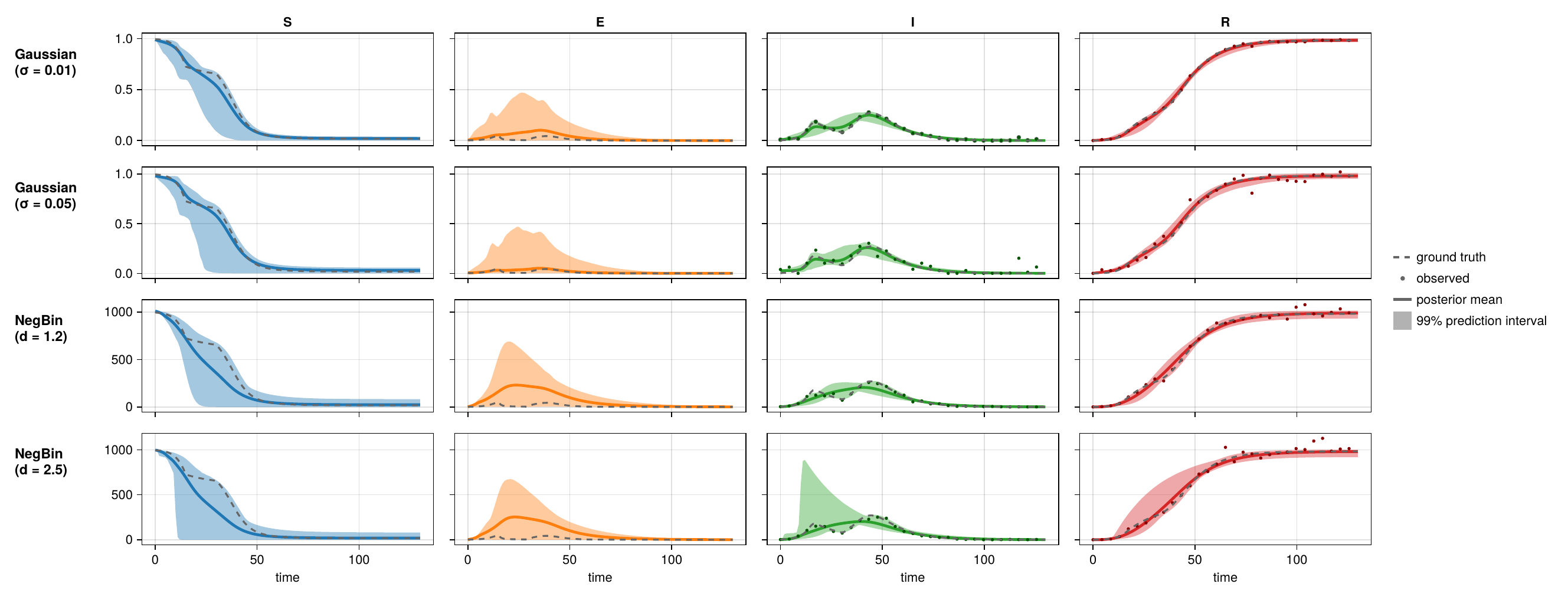}}
\caption{Comparison of different noise model scenarios for the SEIR Pulse problem when using the ensemble-based UQ method.}
\label{fig:seir_pulse_noise_model}
\end{center}
\vskip -0.2in
\end{figure*}

\begin{figure*}[ht]
\vskip 0.2in
\begin{center}
\centerline{\includegraphics[width=0.6\textwidth]{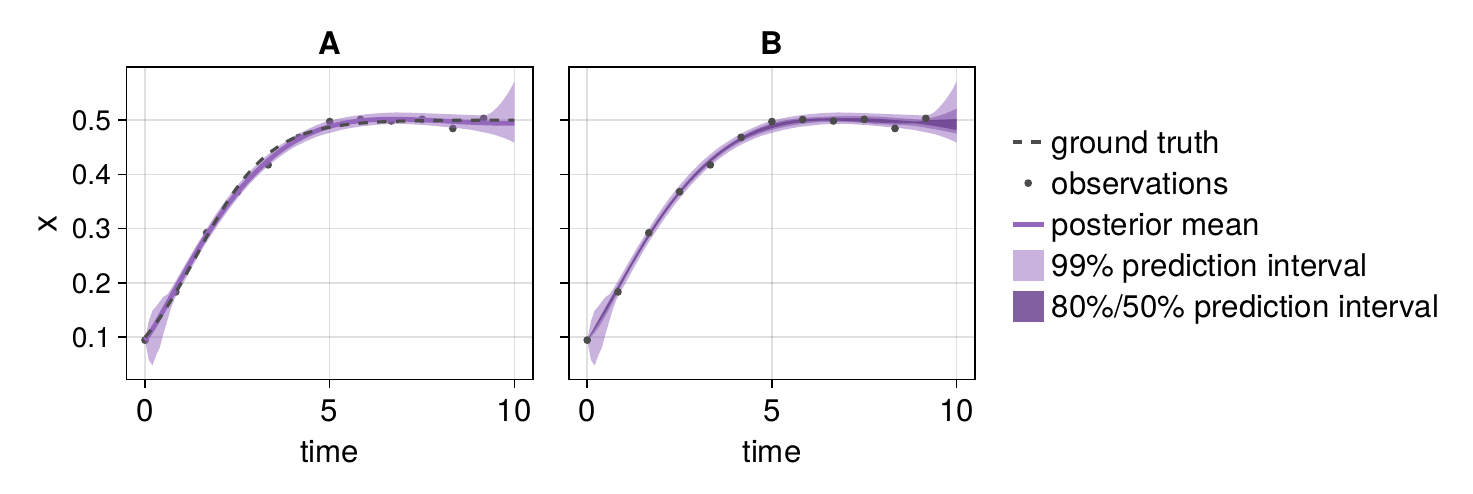}}
\caption{Visualisation of the ensemble-based UQ method for the Quadratic Dynamics problem (Gaussian noise, $\sigma=0.01$)}
\label{fig:quadratic_dynamics}
\end{center}
\vskip -0.2in
\end{figure*}

\clearpage
\subsection{Additional figures for UQ based on MCMC}

\begin{figure*}[ht]
\vskip 0.2in
\begin{center}
\centerline{\includegraphics[width=1\textwidth]{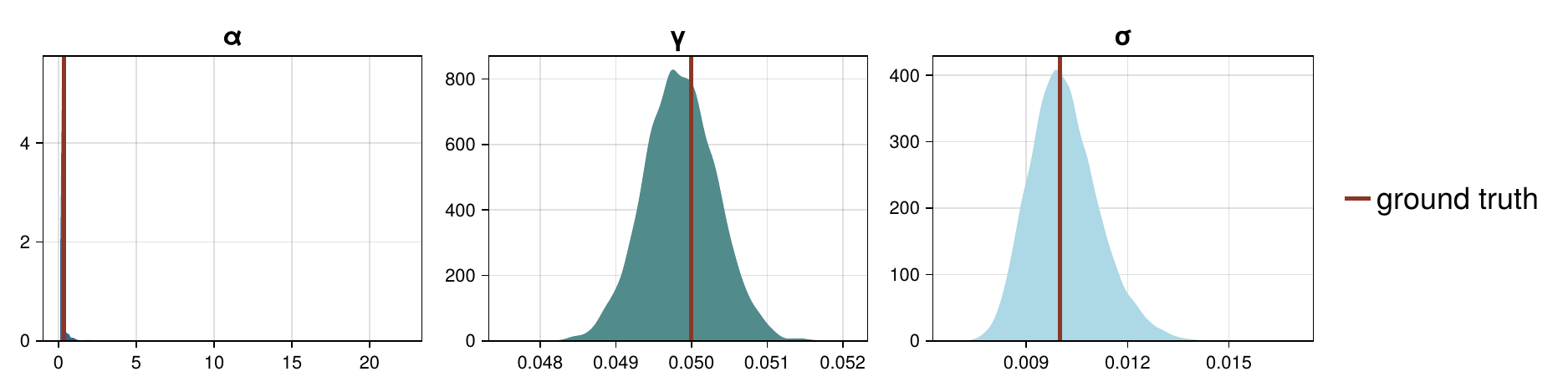}}
\caption{Visualization of the MCMC-based uncertainty quantification for the constant mechanistic parameters (of both the dynamic and noise model) of the SEIR Waves scenario with Gaussian noise (0.01). The red line indicates what parameter value was used to generate synthetic data.}
\label{fig:constant_parameters_MCMC}
\end{center}
\vskip -0.2in
\end{figure*}

\begin{figure*}[ht]
\vskip 0.2in
\begin{center}
\centerline{\includegraphics[width=0.6\textwidth]{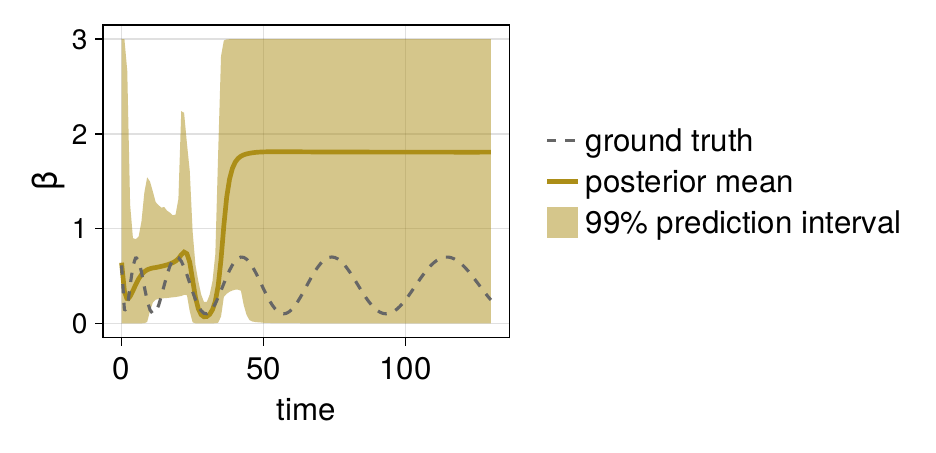}}
\caption{Visualization of the MCMC-based uncertainty of the neural network prediction of $\beta$ for the SEIR Waves scenario with Gaussian noise (0.01).}
\end{center}
\vskip -0.2in
\end{figure*}

\clearpage
\subsection{Additional figures for UQ based on Variational Inference}

\begin{figure*}[ht]
\vskip 0.2in
\begin{center}
\centerline{\includegraphics[width=1\textwidth]{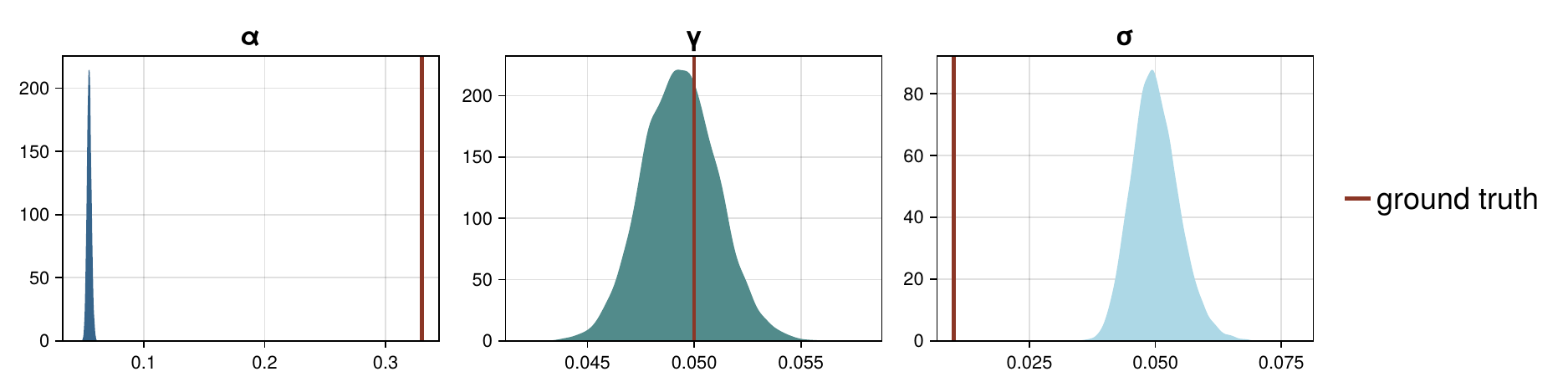}}
\caption{Visualization of UQ for the constant mechanistic parameters (of both the dynamic and noise model) using Variational Inference of the SEIR Waves scenario with Gaussian noise (0.01). The red line indicates what parameter value was used to generate synthetic data.}
\label{fig:constant_parameters_vi}
\end{center}
\vskip -0.2in
\end{figure*}

\begin{figure*}[ht]
\vskip 0.2in
\begin{center}
\centerline{\includegraphics[width=0.6\textwidth]{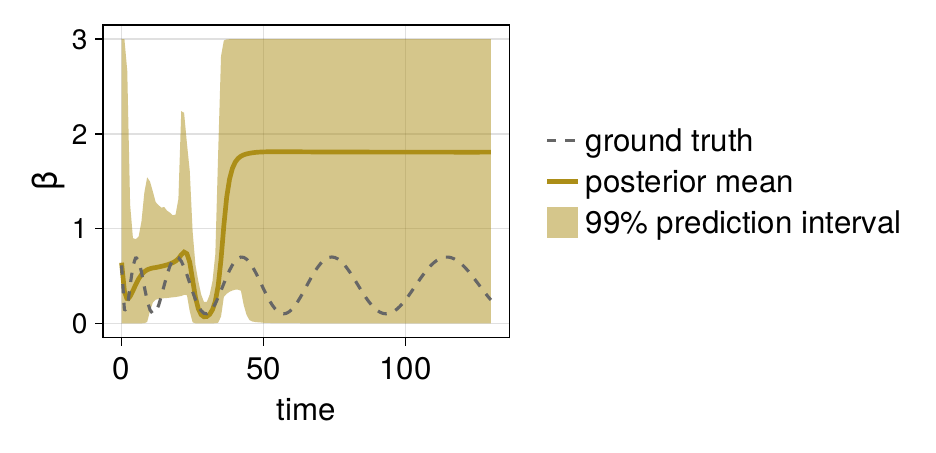}}
\caption{Visualization of UQ of the neural network prediction of $\beta$ for the SEIR Waves scenario with Gaussian noise (0.01) using Variational Inference.}
\end{center}
\vskip -0.2in
\end{figure*}
\clearpage

\subsection{Additional figures for the method comparison}
\begin{figure*}[ht]
\vskip 0.2in
\begin{center}
\centerline{\includegraphics[width=1\textwidth]{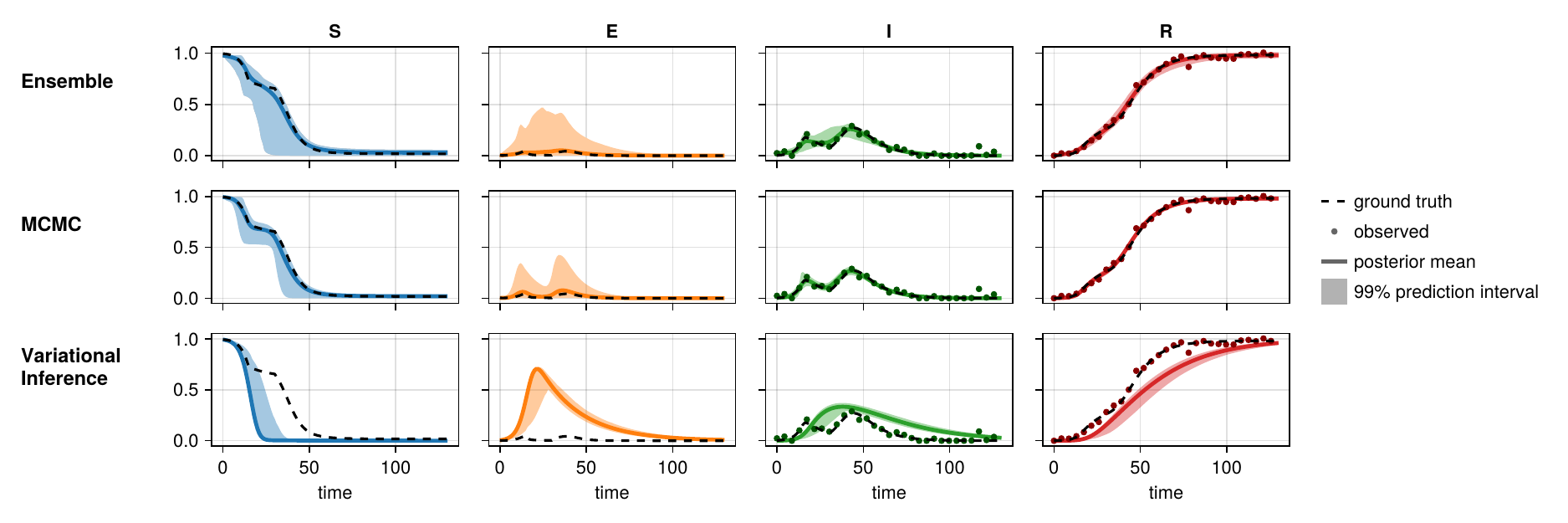}}
\caption{Comparison of the UQ methods on the SEIR Pulse problem (Gaussian noise, $\sigma=0.03$).}
\label{fig:method_comparison_pulse_003}
\end{center}
\vskip -0.2in
\end{figure*}

\clearpage
\newpage

\section{UMAP analysis}
\label{apdx:UMAP}
\counterwithin*{figure}{section}
\renewcommand{\thefigure}{S\arabic{figure}}
\setcounter{figure}{13}
We performed a Uniform Manifold Approximation and Projection (UMAP) analysis \cite{mcinnes2020umap} on the posterior parameter samples obtained from the MCMC and ensemble-based UQ methods. For \cref{fig:umap}, we used the hyperparameters n\_neighbors=10, min\_dist=0.1 and n\_epochs=200. However, the general trend of what is observed was robust to the choice of UMAP hyperparameters.

\begin{figure}[h]
\vskip 0.2in
\begin{center}
\centerline{\includegraphics[width=0.9\columnwidth]{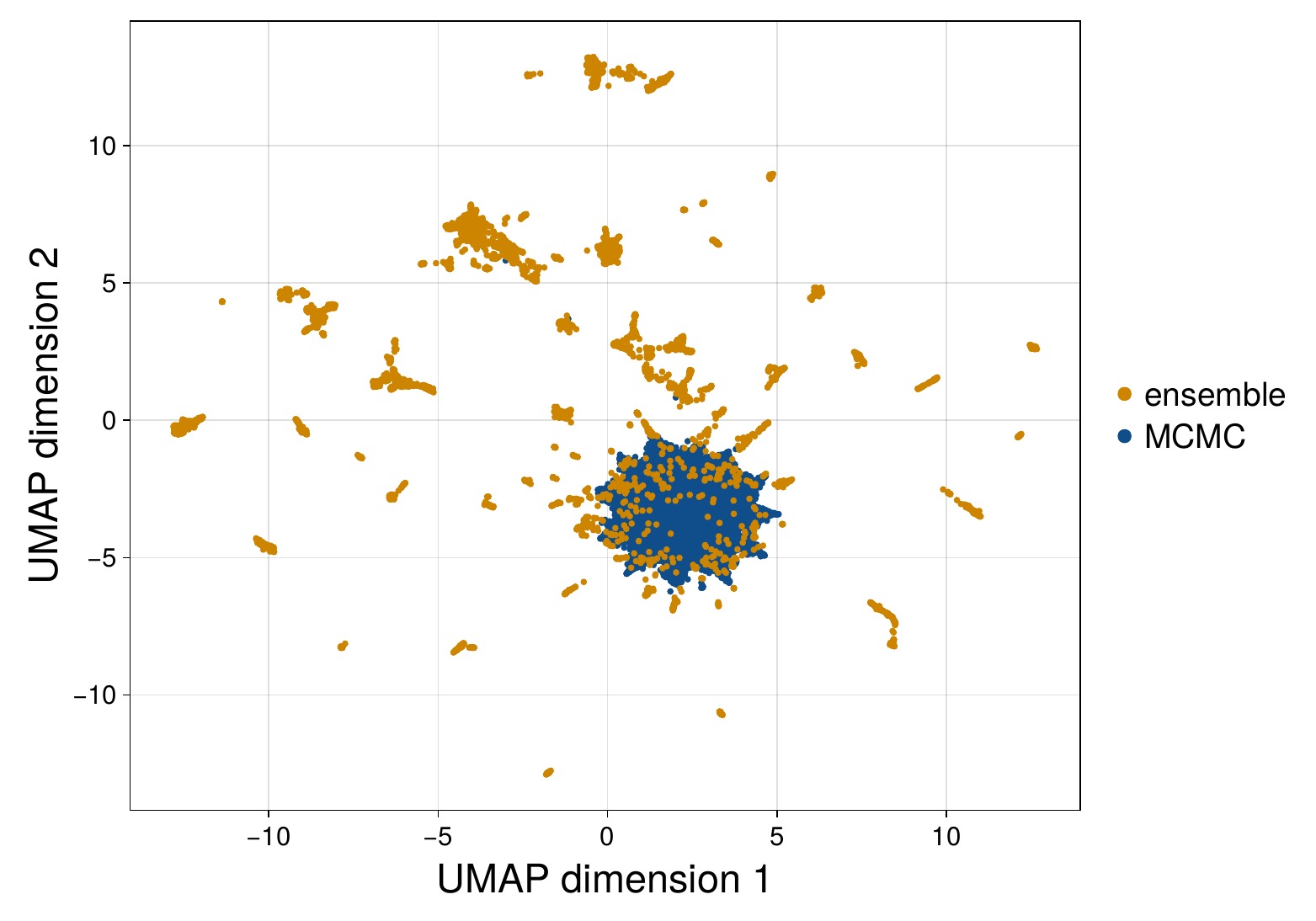}}
\caption{UMAP for the set of parameters obtained by the MCMC and ensemble-based method for the SEIR Waves problem with Gaussian noise (0.01).}
\label{fig:umap}
\end{center}
\vskip -0.2in
\end{figure}

%%%%%%%%%%%%%%%%%%%%%%%%%%%%%%%%%%%%%%%%%%%%%%%%%%%%%%%%%%%%%%%%%%%%%%%%%%%%%%%
%%%%%%%%%%%%%%%%%%%%%%%%%%%%%%%%%%%%%%%%%%%%%%%%%%%%%%%%%%%%%%%%%%%%%%%%%%%%%%%

\end{document}